\pdfoutput=1

\documentclass[11pt]{article}

\usepackage{acl}

\usepackage{times}
\usepackage{latexsym}

\usepackage{footnote}

\usepackage[T1]{fontenc}

\usepackage[utf8]{inputenc}

\usepackage{microtype}
\usepackage{algorithm}
\usepackage[noend]{algpseudocode}
\usepackage{mathtools}
\usepackage{paralist}
\usepackage{hyperref}
\usepackage{booktabs}
\usepackage{enumitem}

\definecolor{ForestGreen}{HTML}{228b22}

\algnewcommand\algorithmicinput{\textbf{Input:}}
\algnewcommand\INPUT{\item[\algorithmicinput]}
\algnewcommand\algorithmicoutput{\textbf{Output:}}
\algnewcommand\OUTPUT{\item[\algorithmicoutput]}

\usepackage{xspace}
\newcommand{\context}{REG-in-context\xspace}
\newcommand{\shot}{one-shot REG\xspace}

\newcommand{\ul}[1]{\underline{#1}}
\newcommand{\bl}[1]{\textbf{#1}}

\title{Non-neural Models Matter: \\A Re-evaluation of Neural Referring Expression Generation Systems}

\author{Fahime Same\textsuperscript{$\heartsuit$}\thanks{\hspace{0.2cm}Equal contribution. Order determined by swapping the order in~\citet{chen-etal-2021-neural-referential}.}, 
Guanyi Chen\textsuperscript{$\spadesuit$}\footnotemark[1], \and
Kees van Deemter\textsuperscript{$\spadesuit$}\\
\textsuperscript{$\heartsuit$}Department of Linguistics, University of Cologne\\
\textsuperscript{$\spadesuit$}Department of Information and Computing Sciences, Utrecht University\\
\texttt{f.same@uni-koeln.de, g.chen@uu.nl, c.j.vandeemter@uu.nl}}

\begin{document}
\maketitle
\begin{abstract}
In recent years, neural models have often outperformed rule-based and classic Machine Learning approaches in NLG. 
These classic approaches are now often disregarded, for example when new neural models are evaluated. 
We argue that they should not be overlooked, since for some tasks, well-designed non-neural approaches achieve better performance than neural ones.
In this paper, the task of generating referring expressions in linguistic context is used as an example. We examined two very different English datasets (\textsc{webnlg} and \textsc{wsj}), and evaluated each algorithm using both automatic and human evaluations.
Overall, the results of these evaluations suggest that rule-based systems with simple rule sets achieve on-par or better performance on both datasets compared to state-of-the-art neural REG systems. In the case of the more realistic dataset, \textsc{wsj}, a machine learning-based system with well-designed linguistic features performed best. We hope that our work can encourage researchers to consider non-neural models in future.
\end{abstract}

\section{Introduction} \label{sec:intro}

Natural Language Generation (NLG) is concerned with the generation of natural language text from non-linguistic input~\citep{10.5555/3241691.3241693}. One step in a classic generation pipeline~\citep{reiter_dale_2000} is Referring Expression Generation (REG, \citet{krahmer2019} for an overview). 
REG has important practical value for commercial natural language generation~\citep{reiter-2017-commercial}, computer vision~\citep{mao2016generation}, and robotics~\citep{fang2015embodied}, for example. 
It has also been used as a tool to understand human language use~\citep{van2016computational}.
REG contains two different problems. One is to find a set of attributes to single out a referent from a set (also called \emph{\shot}). 
The other is to generate referring expressions (REs) to refer to a referent at different points in a discourse~\citep{belz2007generation}. We will focus on the latter task. We call this the \emph{\context} task.

In earlier works, REG is often tackled in two steps \citep{henschel2000pronominalization, krahmer2002efficient}. 
The first step decides the form of an RE. For example, whether a reference should be a proper name (``\emph{Marie Skłodowska-Curie}''), a description (``\emph{the physicist}''), or a pronoun (``\emph{she}'') at a given point in the context.
The second step is concerned with content selection, i.e., the different ways in which a referential form can be realised. 
For example, to generate a description of \emph{Marie Curie}, the REG system  decides whether it is sufficient to mention her profession (i.e., ``\emph{the physicist}'') or whether it is better to mention her nationality as well (i.e., ``\emph{a Polish-French physicist}'').

Thanks to the rapid development of deep learning techniques, recent NLG models are able to generate RE in an End2End (E2E) manner, i.e., to tackle the selection of form and content simultaneously~\citep{castro-ferreira-etal-2018-neuralreg, cao-cheung-2019-referring, cunha-etal-2020-referring}. 
The task of End2End (E2E) REG was proposed by~\citet{castro-ferreira-etal-2018-neuralreg}, who extracted a corresponding corpus from the WebNLG corpus~\citep{castro-ferreira-etal-2018-enriching}\footnote{We refer to this extracted REG corpus as \textsc{webnlg}.}.
Grounding on the \textsc{webnlg} dataset, they proposed a neural REG system built on a {\em sequence-to-sequence with attention} model. Their automatic and human evaluation results suggested that neural REG systems significantly outperform rule-based and feature-based machine learning (ML) baselines.
However, it can be argued that \citeauthor{castro-ferreira-etal-2018-neuralreg} did not use very strong baselines for their comparison: \texttt{OnlyName} is a rule-based system that always generates a proper name given an entity, and \texttt{Ferreira} is a feature-based model that uses Naive Bayes with only 3 simple features\footnote{The human evaluation in \citet{cunha-etal-2020-referring} showed a slightly different result: the \texttt{OnlyName} model performed as well as the Neural REG models in terms of fluency, grammaticality, and adequacy. However, since their human evaluation involved only two subjects, these outcomes need to be approached with caution.}.

\begin{table*}[t!]
\small
\centering
\begin{tabular}{p{15cm}}
\toprule
\textbf{Triples}: \\
(AWH\_Engineering\_College, country, India)\\ 
(Kerala, leaderName, Kochi)\\
(AWH\_Engineering\_College, academicStaffSize, 250)\\
(AWH\_Engineering\_College, state, Kerala)\\
(AWH\_Engineering\_College, city, ``Kuttikkattoor'')\\
(India, river, Ganges) \\
\midrule
\textbf{Text}: AWH Engineering College is in Kuttikkattoor, India in the state of Kerala. The school has 250 employees and Kerala is ruled by Kochi. The Ganges River is also found in India. \\ \midrule
\textbf{Delexicialised Text}: \\
\textbf{Pre-context}: \underline{AWH\_Engineering\_College} is in \underline{``Kuttikkattoor''} , \underline{India} in the state of \underline{Kerala} .\\
\textbf{Target Entity}: \underline{AWH\_Engineering\_College} \\
\textbf{Post-context}: has 250 employees and \underline{Kerala} is ruled by \underline{Kochi} . The \underline{Ganges} River is also found in \underline{India} .\\
\bottomrule
\end{tabular}
\caption{An example data from the \textsc{webnlg} corpus. In the delexicalised text, every entity is \underline{underlined}.}
\label{tab:sample}
\end{table*}

We present several rule-based and feature-based baselines to examine how neural models perform against ``well-designed'' non-neural alternatives. 
Note that a well-designed model is not necessarily complex.
For example, it can be a rule-based system with one or two simple, ``well-designed'' rules.
Since one of the advantages of neural E2E models is that they require little effort for feature engineering, we used two types of baselines, namely models that require minimal expert effort and models that use more demanding (but linguistically well-established) rules or features. Therefore, our main research question is: \emph{Do state-of-the-art neural REG models always perform better than rule-based and machine learning-based models?}

To answer this question fairly, we consider the amount of resources used by each model. For example, the neural models require fewer human resources when it comes to linguistic expertise and annotation, but they require input from Deep Learning experts. Resources such as computing power and data needs should also be considered.

Another issue with previous studies concerns the datasets that were used: in \textsc{webnlg}, approximately 99.34\% of entities in the test set also appear in the training set; consequently, evaluations using \textsc{webnlg} do not take unseen entities into consideration. Furthermore, since many sentences in \textsc{webnlg} are paraphrases of one another, evaluating neural models on \textsc{webnlg} alone may overestimate their performance. \citet{castro-ferreira-etal-2019-neural} recently extended \textsc{webnlg} to include unseen domains that contain many unseen entities\footnote{We used version 1.5 of the \textsc{webnlg} dataset in \url{https://github.com/ThiagoCF05/webnlg}.}, and \citet{cunha-etal-2020-referring} have developed new models to handle them. Their test set has two subsets: one consists of documents 99.34\% of whose entities are {\em seen}, while the other consists of documents 92.81\% of whose entities are {\em unseen}. This arguably makes the data in \textsc{webnlg} unrealistic (see \S\ref{sec:dataset} for discussion).
Therefore, we created what we believe to be a more realistic dataset based on the Wall Street Journal (\textsc{wsj}) portion of the OntoNotes corpus~\citep{hovy-etal-2006-ontonotes,Weischedel2013}\footnote{We used Ontonotes 5.0 licensed by the Linguistic Data Consortium (LDC) \url{https://catalog.ldc.upenn.edu/LDC2013T19}.}.

We evaluate all models on both \textsc{webnlg} and \textsc{wsj}, using automatic and human evaluation experiments. The human experiments included a total of 240 participants and 16920 judgments.

This paper is structured as follows: in \S\ref{sec:dataset} and \S\ref{sec:model}, we describe the datasets used and the REG models. In \S\ref{sec:evaluation}, we provide a detailed description of our automatic and human evaluations. In \S\ref{sec:discussion} and \S\ref{sec:conclusion}, we compare the results across different dimensions and make suggestions for future studies. The code for reproducing the results in this article can be found at: \url{https://github.com/a-quei/neuralreg-re-evaluation}.

\section{Task and Datasets} \label{sec:dataset}
This section explains the \context task and the two English datasets used to conduct the experiments.

\subsection{The \context Task}

Given a text whose REs have not yet been generated, and given the intended referent for each of these REs, the \context task is to build an algorithm that generates all these REs.


Consider the delexicalised text in Table~\ref{tab:sample}. Given the entity ``\emph{AWH\_Engineering\_College}'', REG selects an RE based on the entity and its pre-context (``\emph{AWH\_Engineering\_College is in ``Kuttikkattoor'' , India in the state of Kerala . }''), and its post-context (``\emph{has 250 employees and Kerala is ruled by Kochi . The Ganges River is also found in India.}'').

\subsection{The \textsc{webnlg} Dataset} \label{sec:webnlg}

\citet{gardent-etal-2017-creating} introduced the \textsc{webnlg} corpus for evaluating NLG systems. Using crowd-sourcing, each crowdworker was asked to write a description for a given Resource Description Framework (RDF) triple (Table~\ref{tab:sample}).
The number of triples varied from 1 to 7. This corpus was later enriched and delexicalised~\citep{castro-ferreira-etal-2018-neuralreg, castro-ferreira-etal-2018-enriching} to fit the \context task.
\citet{castro-ferreira-etal-2019-neural} further extended \textsc{webnlg} and divided the documents into test sets \emph{seen} (where all data are from the same domains as the training data) and \emph{unseen} (where all data are from different domains than the training data).
This results in that almost all entities from the seen test set appear in the training set (9580 out of 9644), while only a few entities from the unseen test set appear in the training set (688 out of 9644).
Note that the maximum number of triples in the unseen set is five. So, one would expect the data in the unseen set to be less complex than the seen data.

We used version 1.5 of \textsc{webnlg}, which contains 67,027, 8278, and 19,210 REs in the training, development, and test sets.
From the point of view of the present study, \textsc{webnlg} has some notable shortcomings. For a start, it consists of rather formal texts that may not reflect the everyday use of REs, and in which very simple syntactic structures dominate. 
The texts in \textsc{webnlg} also stand out for other reasons. For example, the texts are extremely short, with an average length of only 1.4 sentences. Consequently, as many as 85\% of the REs are first-mentions, while 71\% of the REs are proper names. 
Finally, in any given test sample, either more than 90\% of the entities are seen or more than 90\% are unseen. 
Realistic data should contain a reasonable amount of mixtures of seen and unseen entities.
For all these reasons, we decided to test all algorithms on a second corpus as well.

\subsection{The \textsc{wsj} Dataset}\label{subsec:wsjdataset}
Using the Wall Street Journal portion of the OntoNotes corpus, we constructed a new English REG dataset, following a similar approach as \citet{castro-ferreira-etal-2018-neuralreg}. This corpus (\textsc{wsj}) has very different characteristics from the \textsc{webnlg}. 
The \textsc{wsj} consists of 582 newspaper articles containing 20,186, 2362 and 2781 REs in the training, development and test sets, respectively. The average length of the documents is 1189 words, and each document consists of 25 sentences on average. Furthermore, 23\% of the instances are first-mention REs and the rest are subsequent mentions.

For each RE, we created its pre- and post-context at the local sentence-level and added $K$ preceding and following sentences to the local context. We refer to $K$ as the context length and set $K$=2 for this experiment. 
To create the dataset, we first delexicalised the REs. The dataset contains nearly 8000 coreferential chains. The REs in each chain were replaced with corresponding delexicalised expressions (similar to table \ref{tab:sample}). 
For delexicalisation, we used (1) the POStag information, (2) the fine-grained annotation of the referential forms, and (3) the entity type of each referent. To delexicalise human REs, for example, we looked for concise but informative REs such as the combination of first and last names (e.g., \emph{``Barack Obama''})\footnote{The reason for choosing concise and informative REs for delexicalisation is that these labels are also used in the realisation process.}. When such an expression was found in a coreferential chain, its delexicalised version (tokens being separated by underscores, e.g., \emph{``Barack\_Obama''}) was assigned to all REs in the chain. We then moved on to the next tag. Below is the order in which the human referents were searched and delexicalised: $[$firstname-lastname$]$, $[$title-firstname-lastname$]$, $[$modified firstname-lastname$]$, $[$title-lastname$]$, $[$lastname$]$, $[$modified-lastname$]$, $[$firstname$]$. For more details on the preparation of the \textsc{wsj} documents and a delexicalised example, see Appendix~\ref{sec:wsjimplementation}.

\section{REG Models} \label{sec:model}

In this section, we introduce the rule-based, ML-based, and the SOTA neural REG models. 
The term \emph{ML-based} here refers to models that require feature engineering and follow a pipeline architecture.

\subsection{Rule-based REG} 

Rule-based models have been widely used for generating REs in context~\citep{ mccoy-strube-1999-generating, henschel2000pronominalization}. 
Here, we build rule-based systems for binary classification into two classes, namely pronominal and non-pronominal REs.

\paragraph{Simple Rule-based System (\texttt{RREG-S}).}
For the first rule-based system, we use 2 simple rules.
The target entity $r$ is realised as a pronominal RE if: 
\begin{enumerate}[noitemsep]
    \item $r$ is \emph{discourse-old};
    \item $r$ has no \emph{competitor} in the current sentence and the previous sentence,
\end{enumerate}
Otherwise, $r$ is realised as a non-pronominal RE.
An entity $r$ is defined as discourse-old if it has been mentioned in the previous context. A competitor is an entity that can be referred to with the same pronoun as $r$. 

We also build a dictionary that stores the pronouns associated with each entity. For seen entities, we extract pronouns from the training data. If an entity has multiple possible pronominal forms, we extract the most frequent one. For unseen entities, we determine their pronominal forms based on their meta-information, which is also used in E2E systems~\citep{cunha-etal-2020-referring}. For example, if an entity in \textsc{webnlg} has the type \texttt{PERSON} and the gender \texttt{FEMALE}, we assign ``\emph{she}'' to this entity.

For the surface realisation of each entity, we realise its non-pronominal form by replacing the underscores in the entity label with whitespaces (e.g., ``\emph{Adenan\_Satem}'' to ``\emph{Adenan Satem}''), as previously described by~\citet{castro-ferreira-etal-2018-neuralreg}. We realise the pronominal forms according to~\citet{castro-ferreira-etal-2016-towards-variation} by using the grammatical role of each entity (e.g., if the entity is in the object position, then we realise ``\emph{he}'' as ``\emph{him}'').

\paragraph{Linguistically-informed Rule-based System (\texttt{RREG-L}).}
We build \texttt{RREG-L} by adopting a set of pronominalisation rules from~\citet{henschel2000pronominalization}.
The fundamental concepts used by these rules are the idea of \emph{local focus}, which is a simpler implementation of the Centering Theory \citep{grosz-etal-1995-centering}, and \emph{parallelism}, i.e., whether $r$ and its antecedent in the previous sentence have the same grammatical role~\citep{henschel2000pronominalization}.
The \texttt{RREG-L} is described in detail in Appendix~\ref{sec:rreg-L}.



\subsection{ML-based REG}
The GREC Shared Task~\citep{belz2010generating} triggered a plethora of ML-based models for building \context \citep[e.g.,][]{greenbacker2009udel,hendrickx-etal-2008-cnts}. These models differ from each other in the features and the ML algorithms they have used.

In this study, we build ML-based REG models using CatBoost~\citep{Prokhorenkova2018}. It predicts whether a reference is realised as a pronoun, proper name, or description. Once the referential form is predicted, the next step is to select the content. The most frequent variant (with the same referential form as the predicted class) is selected in the training corpus given the referent and the full set of features. If no matching RE is found, a back-off method \citep{castro-ferreira-etal-2018-neuralreg} is used, removing one feature at a time in order of importance. The order is calculated using the inherent feature importance method of the CatBoost algorithm. Depending on which features are used, we build two variants of ML-based models, namely \texttt{ML-S} and \texttt{ML-L}. The detailed list of the features used in these models can be found in Appendix~\ref{sec:feature}.

\paragraph{Features obtained by minimum effort (\texttt{ML-S}).}
To find out what the upper bound is for a system that does not require any additional linguistic information or any additional annotation effort, we developed \texttt{ML-S}. In this model, we have relied only on the features that can be extracted directly from the corpus. Therefore, features such as grammatical role (which requires a syntactic parser) are not included in this model.

\paragraph{Linguistically Informed Features (\texttt{ML-L}).}
To evaluate the upper bound performance of ML-based systems, we developed \texttt{ML-L} with the features that could affect the choice of referential form and could improve the overall accuracy of the REG systems suggested by the previous linguistic and computational studies \citep{ariel1990accessing,gundel1993cognitive,brennan1995centering,arnold2007effect,fukumura2011effect,kibrik2016referential,von2019discourse,same-van-deemter-2020-linguistic}. For example, we included features encoding grammatical role, recency, gender, and animacy in \texttt{ML-L}.
Note that \texttt{ML-L} makes full use of the syntactic information\footnote{The syntactic information is available in the annotations of the \textsc{wsj} data set. We used spaCy to parse \textsc{webnlg}.} and entity meta-information (e.g., \texttt{GENDER} and \texttt{TYPE} which are also used by both the rule-based systems and the neural models).

\begin{table*}[t]
\small
\centering
\begin{tabular}{l|cccc|ccc}
\toprule
Model & RE Acc.$\uparrow$ & SED$\downarrow$ & BLEU$\uparrow$ & Text Acc.$\uparrow$ & Precision$\uparrow$ & Recall$\uparrow$ & F1$\uparrow$ \\ \midrule
\texttt{RREG-S} & \underline{54.60} & \textbf{3.65} & \underline{72.05} & 16.28 & \textbf{89.52} & \underline{77.57} & \underline{82.28} \\
\texttt{RREG-L} & 53.43 & 3.77 & 71.27 & 15.49 & 73.94 & 73.96 & 73.95 \\
\texttt{ML-S} & 54.35 & 3.70 & 70.89 & 15.43 & 71.70 & 63.52 & 66.39 \\
\texttt{ML-L} & \textbf{56.69} & \underline{3.66} & \textbf{72.25} & \underline{16.36} & 81.66 & 63.62 & 68.36 \\
\texttt{ATT+Copy} & 48.75 & 4.46 & 68.48 & 14.88 & 85.33 & 75.74 & 79.63 \\
\texttt{ATT+Meta} & 53.34 & 4.22 & 70.82 & \textbf{16.54} & 86.32 & 75.56 & 79.81 \\
\texttt{ProfileREG} & 40.96 & 7.40 & 61.04 & 11.39 & \underline{86.44} & \textbf{87.40} & \textbf{86.91} \\
\bottomrule
\end{tabular}
\caption{Automatic Evaluation Results on \textsc{webnlg}. Best results are \textbf{boldfaced}, whereas the second best are \underline{underlined}. $\uparrow$ means the higher the metric, the better, while $\downarrow$ means the lower the better.}
\label{tab:webnlg_result}
\end{table*}

\subsection{Neural REG}
A limitation of the rule-based and ML-based models mentioned above is that they are not able to handle situations where an RE form (e.g., a proper name) can have multiple realisations, e.g., Lady Gaga/Stefani Germanotta. End2End NeuralREG can address this by generating REs from scratch.
This study examines three NeuralREG systems that have been developed to deal with unseen entities as well. All of them were developed using the {\em sequence-to-sequence with attention} model~\citep{bahdanau2014neural}.

\paragraph{\texttt{ATT+Copy}.}
\citet{cunha-etal-2020-referring} proposed using three bidirectional LSTMs \citep{hochreiter1997long} to encode a pre-context, a post-context, and the proper name of an entity (i.e., replacing underscores in entity labels with whitespaces) into three hidden vectors $h^{(pre)}$, $h^{(post)}$ and $h^{(r)}$, respectively.
An auto-regressive LSTM-based decoder
generates REs based on context vectors.
To handle unseen entities, \citeauthor{cunha-etal-2020-referring} used the copy mechanism, which allows the decoder to copy words from the contexts directly as output.

\paragraph{\texttt{ATT+Meta}.} \texttt{ATT+Meta}~\citep{cunha-etal-2020-referring} used meta information of each entity to improve the quality of the generated REs. 
In each decoding step $t$, the context vector $\mathbf{v}^{(c)}_t$ is concatenated with meta information embeddings before being fed to the decoder. In \textsc{webnlg}, meta information are the entity type $\textbf{v}^{(type)}$ and gender embeddings $\textbf{v}^{(gender)}$; while in \textsc{wsj}, in addition to $\textbf{v}^{(type)}$ and $\textbf{v}^{(gender)}$, there is also plurality embedding $\textbf{v}^{(pl)}$.

\paragraph{\texttt{ProfileREG}.}
\citet{cao-cheung-2019-referring} made \texttt{ProfileREG} to leverage the content of entity profiles extracted from Wikipedia. More specifically, instead of encoding the proper name of each entity, \texttt{ProfileREG} asks the entity encoder to encode the whole entity's profile to obtain $h^{(r)}$.
Note that since profiles of entities in \textsc{wsj} are not accessible, we evaluate \texttt{ProfileREG} only on \textsc{webnlg}.

\section{Evaluation}\label{sec:evaluation}

\begin{table*}[t]
\small
\begin{tabular}{l|cccc|ccc}
\toprule
Model & RE Acc.$\uparrow$ & SED$\downarrow$ & BLEU$\uparrow$ & Text Acc.$\uparrow$ & Precision$\uparrow$ & Recall$\uparrow$ & F1$\uparrow$ \\ \midrule
\texttt{RREG-S} & 54.76/\textbf{54.44} & 3.94/\textbf{3.35} & 73.98/\textbf{69.90} & 18.85/\textbf{13.56} & 85.50/\textbf{91.51} & 71.21/\underline{81.61} & 76.29/\textbf{85.73} \\
\texttt{RREG-L} & 54.10/\underline{52.75} & 3.86/\underline{3.68} & 73.56/\underline{68.72} & 18.49/\underline{12.41} & 69.66/77.65 & 73.34/74.43 & 71.30/75.90 \\
\texttt{ML-S} & 58.61/50.06 & 3.38/4.02 & 73.73/67.67 & 18.65/12.12 & 74.15/56.99 & 85.05/50.26 & 78.35/48.24 \\
\texttt{ML-L} & 63.01/50.32 & 3.30/4.03 & 76.10/67.91 & 20.30/12.33 & 83.68/73.90 & 85.00/50.21 & 84.32/47.98 \\
\texttt{ATT-Copy} & \underline{71.47}/25.84 & \underline{2.64}/6.28 & \textbf{80.46}/54.50 & \underline{26.39}/3.08 & \textbf{86.90}/83.66 & 87.75/68.12 & \underline{87.32}/72.97 \\
\texttt{ATT-Meta} & \textbf{71.64}/34.88 & \textbf{2.62}/5.82 & \underline{80.26}/60.00 & \textbf{27.88}/4.93 & 86.48/\underline{86.66} & \underline{87.97}/67.72 & 87.21/73.14 \\
\texttt{ProfileREG} & 68.26/13.43 & 3.27/11.55 & 78.24/41.13 & 21.82/0.7 & \underline{86.79}/86.04 & \textbf{94.80}/\textbf{82.66}  & \textbf{90.33}/\underline{84.24} \\
\bottomrule
\end{tabular}
\caption{Automatic Evaluation Results of \textsc{webnlg} for seen/unseen data respectively.}
\label{tab:seen_unseen_result}
\end{table*}

\begin{table*}[t]
\small
\centering
\begin{tabular}{l|cccc|ccc}
\toprule
Model & RE Acc.$\uparrow$ & SED$\downarrow$ & BLEU$\uparrow$ & Text Acc.$\uparrow$ & Precision$\uparrow$ & Recall$\uparrow$ & F1$\uparrow$ \\ \midrule
\texttt{RREG-S} & 35.89 & 12.54 & 81.71 & 34.78 & 56.34 & 53.00 & 51.73 \\
\texttt{RREG-L} & \underline{37.22} & 12.37 & 82.06 & 36.07 & 67.11 & 54.31 & 52.08 \\
\texttt{ML-S} & 37.18 & 12.56 & \underline{82.28} & 36.07 & \underline{77.93} & 56.70 & 55.52 \\
\texttt{ML-L} & \textbf{56.60} & \textbf{9.23} & \textbf{85.64} & \textbf{50.03} & \textbf{85.12} & \textbf{85.75} & \textbf{85.43} \\
\texttt{ATT+Copy} & 29.27 & 15.19 & 79.01 & 32.55 & 76.33 & 54.16 & 51.10 \\
\texttt{ATT+Meta} & 35.56 & \underline{12.11} & 81.07 & \underline{36.83} & 72.72 & \underline{70.50} & \underline{71.42} \\
\bottomrule
\end{tabular}
\caption{Automatic Evaluation Results on \textsc{wsj}. Note that since, for a \textsc{wsj} document, it is extremely unlikely to generate every RE correctly, the text accuracy is always 0. Instead, we report sentence-level accuracy. 
}
\label{tab:wsj_filename}
\end{table*}

We evaluated all the systems described in \S\ref{sec:model} on both \textsc{webnlg} and \textsc{wsj} using automatic and human evaluations. 
We implemented the neural models based on the code of \citet{cunha-etal-2020-referring} and \citet{cao-cheung-2019-referring}\footnote{\texttt{ATT+Copy} and \texttt{ATT+Meta}: \url{github.com/rossanacunha/NeuralREG}; and \texttt{ProfileREG}: \url{github.com/mcao610/ProfileREG}.}. For \textsc{webnlg}, we used their original parameter setting, while for \textsc{wsj}, we tuned the parameters on the development set and used the best parameter set.

To determine the optimal context length $K$ of \textsc{wsj}, we varied $K$ from 1 to 5 sentences before and after the target sentence, then tested \texttt{ATT+Meta} on the development set with the different $K$ contexts. It reaches the best performance when $K=2$.

\subsection{Automatic Evaluation}

\paragraph{Metrics.} Following~\citet{cunha-etal-2020-referring}, we evaluated REG systems from 3 angles. 
(1) \textbf{RE Accuracy} and \textbf{String Edit Distance}~\citep[SED,][]{levenshtein1966binary} were used to evaluate the quality of each generated RE. (2) After adding the REs to the original document, \textbf{BLEU}~\citep{papineni2002bleu} and \textbf{Text Accuracy} were used to evaluate the output text. (3) \textbf{Precision}, \textbf{recall}, and \textbf{F1} score were used to assess pronominalisation.

\paragraph{Results of \textsc{webnlg}.}

Table~\ref{tab:webnlg_result} depicts the results of \textsc{webnlg}\footnote{Note that there is a discrepancy between our replication results and the results of \citet{cunha-etal-2020-referring}. The reason for this difference is that we found a bug in the code for pre-processing provided by the original paper and fixed it after consultation with \citeauthor{cunha-etal-2020-referring}}.
Overall, the classic rule- and ML-based models performed better than neural models, while neural models did a better job on pronominalisation.
For generating REs, \texttt{ML-L} had the best performance, as it obtained the highest RE accuracy and BLEU scores and the second best SED and text accuracy score.
For pronominalisation, \texttt{ProfileREG} yields the best performance, followed by \textsc{RREG-S}. 

We were surprised to find that the simplest rule-based system, \texttt{RREG-S}, performs remarkably well. 
It not only defeats the linguistically informed, rule-based \texttt{RREG-L}, but also outperforms the SOTA neural models \texttt{ATT+Copy} and \texttt{ATT+Meta} on both RE generation and pronominalisation.

Table~\ref{tab:seen_unseen_result} shows the breakdown of the seen and unseen subsets. The SOTA neural models (i.e., \texttt{ATT+Copy}, \texttt{ATT+Meta}, and \texttt{ProfileREG}) have the top 3 performance on seen data, and the worst RE generation performance (i.e., RE Acc., SED, BLEU, and Text Acc.) on unseen data. The ML-based models achieve the fourth and fifth best performance on seen data, and lower performance (but not as low as the neural models) on unseen data.
The nature of \textsc{webnlg} could explain this drop in performance on unseen data: the models may have limited ability to handle unseen entities, for instance, because they fail to conduct domain transfer (remember that unseen data comes from different domains than seen data).
Since rule-based systems do not rely on training data, this explanation does not apply to them, which explains why they did not show the same drop in performance. In fact,
they performed even better on unseen data, possibly because unseen data contained fewer triples than seen data (see \S\ref{sec:dataset}).
Concretely, rule-based systems have lower REG accuracy but higher pronominalisation accuracy on unseen data compared to seen data. Additionally, ML-based models have low performance in the pronominalisation of unseen entities. The pronominalisation accuracy of the rule-based models is based on a 2-way distinction between a pronominal and a non-pronominal form, while the ML-based models make a 3-way distinction between a pronoun, a proper name and a description.

Another factor that might have lowered the performance of the ML models is the annotation practices in \textsc{webnlg}. Since these models are data-driven, the quality of the annotations directly affects their performance. It appears that whenever a (nominal) RE starts with a determiner, it is marked in \textsc{webnlg} as \texttt{description}; otherwise, it is marked as \texttt{proper name}. For instance, \emph{``United States''} is marked as a proper name, while \emph{``The United States''} is wrongly marked as a description. To allow comparison with previous work, we have not corrected the annotations,
but it is important to keep in mind that this issue can cause ML-based models to underperform.

\begin{table*}[t!]
\small
\centering
\begin{tabular}{lccccccccc}
\toprule
 & \multicolumn{3}{c}{All} & \multicolumn{3}{c}{Seen} & \multicolumn{3}{c}{Unseen} \\\cmidrule(lr){2-4} \cmidrule(lr){5-7}\cmidrule(lr){8-10}
Model & Fluency & Grammar & Clarity & Fluency & Grammar & Clarity & Fluency & Grammar & Clarity \\ \midrule
\texttt{RREG-S} & \underline{5.76}$^{A}$ & \textbf{5.73}$^{A}$ & \textbf{5.71}$^{A}$ & \underline{5.73}$^{A}$ & 5.62$^{A}$ & 5.68$^{A}$ & \underline{5.79}$^{A}$ & \textbf{5.83}$^{A}$~~~ & \textbf{5.75}$^{A}$ \\
\texttt{RREG-L} & 5.68$^{A}$ & 5.52$^{A}$ & 5.67$^{A}$ & 5.63$^{A}$ & 5.45$^{A}$ & 5.64$^{A}$ & 5.74$^{A}$ & 5.60$^{A,B}$ & 5.70$^{A}$ \\
\texttt{ML-S} & 5.73$^{A}$ & \underline{5.65}$^{A}$ & \textbf{5.71}$^{A}$ & 5.65$^{A}$ & \underline{5.64}$^{A}$ & 5.70$^{A}$ & \textbf{5.82}$^{A}$ & \underline{5.65}$^{A,B}$ & \underline{5.72}$^{A}$ \\
\texttt{ML-L} & \textbf{5.78}$^{A}$ & 5.63$^{A}$ & 5.67$^{A}$ & 5.73$^{A}$ & 5.63$^{A}$ & 5.62$^{A}$ & \textbf{5.82}$^{A}$ & 5.62$^{A,B}$ & \underline{5.72}$^{A}$ \\
\texttt{ATT+Copy} & 5.65$^{A}$ & 5.62$^{A}$ & \underline{5.68}$^{A}$ & 5.71$^{A}$ & \underline{5.64}$^{A}$ & \underline{5.76}$^{A}$ & 5.58$^{A}$ & 5.60$^{A,B}$ & 5.59$^{A}$ \\
\texttt{ATT+Meta} & 5.68$^{A}$ & 5.56$^{A}$ & 5.66$^{A}$ & 5.69$^{A}$ & \textbf{5.68}$^{A}$ & 5.65$^{A}$ & 5.67$^{A}$ & 5.43$^{B}$~~~ & 5.66$^{A}$ \\
\texttt{ProfileREG} & 5.70$^{A}$ & 5.56$^{A}$ & 5.61$^{A}$ & \textbf{5.81}$^{A}$ & \textbf{5.68}$^{A}$ & \textbf{5.77}$^{A}$ & 5.58$^{A}$ & 5.43$^{B}$~~~ & 5.44$^{A}$ \\ \midrule
\texttt{Human} & 5.81$^{A}$ & 5.69$^{A}$ & 5.82$^{A}$ & 5.77$^{A}$ & 5.69$^{A}$ & 5.83$^{A}$ & 5.84$^{A}$ & 5.70$^{A}$~~~ & 5.80$^{A}$ \\
\bottomrule
\end{tabular}
\caption{Human Evaluation Results on \textsc{webnlg} corpus. Rankings are determined by significance testing ($p < 0.01$). 
Per column, results that have \emph{no} superscript letters in common are significantly different from each other.
}
\label{tab:human_webnlg}
\end{table*}

\begin{table}[t]
\small
\centering
\begin{tabular}{lccc}
\toprule
Model & Fluency & Grammar & Clarity \\ \midrule
\texttt{RREG-S} & 76.13$^{A,B}$ & 75.74$^{A,B,C}$ & 78.03$^{A}$ \\
\texttt{RREG-L} & 72.56$^{A,B}$ & 73.38$^{C}$~~~~~~~ & 74.76$^{A}$ \\
\texttt{ML-S} & 77.48$^{A}$~~~ & 78.39$^{A,B}$~~~~ & 78.76$^{A}$ \\
\texttt{ML-L} & 77.52$^{A}$~~~ & 78.45$^{A}$~~~~~~~ & 79.45$^{A}$ \\
\texttt{ATT+Copy} & 74.43$^{A,B}$ & 74.57$^{A,B,C}$ & 75.63$^{A}$ \\
\texttt{ATT+Meta} & 71.94$^{B}$~~~ & 72.95$^{B,C}$~~~~ & 73.95$^{A}$ \\
\bottomrule
\end{tabular}
\caption{Human Evaluation Results on \textsc{wsj}.} 
\label{tab:human_wsj}
\end{table}

\paragraph{Results of \textsc{wsj}.}

Table~\ref{tab:wsj_filename} shows the results of \textsc{wsj}. Once again, \texttt{ML-L} performs best both in RE generation and in pronominalisation, outperforming the other models by a large margin.
\texttt{RREG-L} outperforms \texttt{RREG-S} on \textsc{wsj} on all evaluation metrics, 
which could be seen as confirmation of our hunch that \textsc{wsj} contains different, and potentially more naturalistic texts than \textsc{webnlg} (see \S\ref{sec:webnlg}). 

As for the neural models, the results suggest that meta-information can improve RE prediction accuracy. Also, the inclusion of meta-information significantly boosts the recall of pronominalisation comparing \texttt{ATT+Copy} with \texttt{ATT+Meta}. Table \ref{tab:generatedwsj} in Appendix \ref{sec:generated_text} shows an original text and different outputs generated by the \textsc{wsj} models.

\subsection{Human Evaluation on \textsc{webnlg}}

\paragraph{Materials.} For \textsc{webnlg} seen entities, we randomly sampled 4 instances from each triple size group of 2-7 from the test set. In the case of the unseen data, we randomly chose 6 instances from size groups of 2-5. In this way, we obtained a total number of 48 reference instances (24 seen and 24 unseen). In addition to each reference instance, we selected its 7 different versions generated by the models (3 neural, 2 ML-based and 2 rule-based models). This yields a total of 384 items (48 $\times$ 8).

\paragraph{Design.} The  384 items were randomly distributed into 12 lists of 32 items. Each list was rated by 10 participants.  Participants were asked to rate each text for its fluency (“does the text flow in a natural, easy to read manner?”), grammaticality (“is the text grammatical (no spelling or grammatical errors)?”) and clarity (“does the text clearly express the data in the table?”) on a 7-point Likert scale anchored by 1 (very bad)
and 7 (very good). The definition of each criterion was taken from \citet{castro-ferreira-etal-2018-neuralreg}. 

\paragraph{Participants.} We used Amazon Mechanical Turk (MTurk) for human evaluation. We restricted MTurk workers to those located in the United States, with an approval rating of $\geq$ 
95\% and 1,000 or more HITs approved. 
We rejected workers if they: 
(1) gave human-produced descriptions a score lower than 2 more than 3 times; or 
(2) gave scores with a standard deviation less than 0.5. 
120 workers (12 lists$\times$10 workers) participated, providing us with 11520 judgements (384 items$\times$3 criteria$\times$10 judgements/item). 
The participants were 80 males, 36 females, and 4 others/unanswered, with an average age of 37.

\paragraph{Results.} Table~\ref{tab:human_webnlg} shows the results of the human evaluation \textsc{webnlg}. Few of the differences reach significance (using Wilcoxon's signed-rank test with Bonferroni correction\footnote{P-values were multiplied by the number of comparisons.}), suggesting that \textsc{webnlg} may be ill-suited for differentiating between REG models\footnote{All non-significant differences in Table~\ref{tab:human_webnlg} and Table~\ref{tab:human_wsj} are associated with p-values greater than $0.1$.}.
The only two significant differences appear when comparing \texttt{RREG-S} with \texttt{ATT+Meta} and \texttt{ProfileREG} in terms of the grammaticality of unseen data.
The results suggest that \texttt{RREG-S} is the best model for generating REs on \textsc{webnlg}, performing on a par with neural models on seen data and better than neural models on unseen data. Unlike our automatic evaluation, \texttt{ATT+Meta} does not outperform \texttt{ATT+Copy} in human evaluation.

\subsection{Human Evaluation on \textsc{wsj}}

\paragraph{Materials.} We randomly selected 30 documents from the test set of \textsc{wsj} (reference text). We included the 6 different outputs generated by the 6 \textsc{wsj} models (hereafter target texts). In this way we obtained a total of 180 reference-target pairs.

\paragraph{Design.} As mentioned in \S\ref{subsec:wsjdataset}, the \textsc{wsj} documents have an average length of 25 sentences.
Since there are no input representations (e.g., in RDF) for \textsc{wsj}, we decided to ask participants to compare texts using a Magnitude Estimation (ME) \citep{bard1996magnitude}.
The participants saw the reference and one of the target texts side by side, and they were asked to rate the target relative to the reference text. To make the task manageable for participants, 
texts were shortened to a maximum of the first 20 sentences. The 180 reference-target pairs were randomly distributed over 12 lists, each list having 15 items. Each list was rated by 10 participants. They were asked to rate the fluency, grammaticality and clarity of the target texts. The definition of fluency and grammaticality were as in the \textsc{webnlg} task, and clarity was defined as ``how clearly does the target text allow you to understand the situation described in the standard text\footnote{We refer to the reference text as \emph{standard text}.}?". The question asked for each of the 3 criteria was: assuming that standard text has a score of 100, how do you rate the fluency$|$grammaticality$|$clarity of target text? Participants were allowed to choose any positive number. 

\paragraph{Participants.} The MTurk worker restrictions were similar to the \textsc{webnlg} experiment. Workers with scores less than 5 standard deviations were rejected. The experiment included 120 participants, resulting in 5400 judgements (180 items$\times$3 criteria$\times$10 judgements/item). The participants were 65 males, 54 females, and 1 others/unanswered, with an average age of 38.

\paragraph{Results.} Since typos are possible in ME (e.g., a worker might type 600 instead of 60), we excluded outliers, defined as a score that is lower than the median 
minus 3 standard deviations, or higher than the median 
plus 3 standard deviations of that item.
The remaining scores were down-sampled for conducting significant testing.
The results are shown in Table~\ref{tab:human_wsj}. 
Unlike \textsc{webnlg}, significant differences are frequent. For fluency, \texttt{ML-S} and \texttt{ML-L} perform the best while \texttt{ATT+Meta} performs the worst.
For grammaticality, \texttt{ML-L} is still the best model, which significantly defeats \texttt{RREG-L} and \texttt{ATT+Meta}.
A more detailed study is needed to investigate why \texttt{RREG-L} is the second worst in terms of grammaticality, which we found surprising.
For clarity, no significant difference was found, perhaps because it was difficult for participants to compare long documents.
In sum, on \textsc{wsj}, \texttt{ML-L} has the best performance, and the simpler \texttt{ML-S} and \texttt{RREG-S} also have considerably good performances. 




\section{Discussion}\label{sec:discussion}



\paragraph{Why does Neural REG not defeat rule-based REG?}

Received wisdom has it that although neural models may be inferior to other models in terms of interpretability, they are nonetheless superior in terms of performance. Although it is possible that future neural models will perform better than the ones examined here, our results call into question whether this received wisdom is correct.
One possible explanation is the observation that Neural NLG systems tend to perform very well on surface realisation tasks, but less well on tasks that focus on semantic content (see e.g.,~\citet{hallucination2018} on hallucinations in the Data2Text generation tasks). REG, after all, is a task that focuses in large part on semantic content.
%
There may be other reasons, which should be investigated in future work.



\paragraph{Role of Linguistically-informed Features.}
Rule-based models did particularly well on \textsc{webnlg}, outperforming other models. By contrast, on \textsc{wsj}, the linguistically-informed feature-based model (\texttt{ML-L}) outperformed all other models. This suggests that the type of text, and consequently, the complexity of the REG task, might be a factor in choosing the REG method. Linguistically-informed features seem to have a more pivotal role in the case of more complex text types, whereas simpler texts can be handled at least as well by simpler rule-based models.

\paragraph{Resources Use.}
As mentioned before, different approaches require different amounts of {\bf human} resources and annotation efforts. But we believe that other resource types should also be taken into consideration when models are compared, including the following:
(1) \textbf{The amount of context}: the neural models access the whole pre-context and post-context for \textsc{webnlg}, while they access $K$ preceding and $K$ following sentences around the target entity for \textsc{wsj}. The ML-based models extract features taking only the current sentence and the whole pre-context into account. The rule-based models only look at the current sentence and the previous one;
(2) \textbf{External tools}: the neural models need no external tools, while the rule-based and \texttt{ML-L} models need a syntactic parser (which is also used for constructing datasets);
(3) \textbf{External information}: rule-based models, \texttt{ML-L}, and \texttt{ATT+Meta} need entities' meta-information. \texttt{ProfileREG} requires the profile description of each entity, which, for most REG tasks, is hard to obtain; (4) \textbf{Computing resources}: the neural models need GPUs while other models can be constructed using merely personal computers;
(5) \textbf{The amount of training data}: the rule-based models need no training data, while other models require training data (large-scale naturalistic versions of which, for the task of REG, is not available).

As we have seen, \texttt{RREG-S} and \texttt{ML-S} perform remarkably well on both \textsc{wsj} and \textsc{webnlg}. Taking resources into consideration, the advantage of using a model such as \texttt{RREG-S} and \texttt{ML-S} becomes more pronounced. \texttt{RREG-S} uses less human resources, less context, less computing resources, and no training data\footnote{The pronominal form (e.g., he or she) of an entity can either be extracted from the training data or decided by the entity's meta-information.} compared to other models. \texttt{ML-S} needs more context and training data; it probably also needs more human effort for feature engineering and selecting ML models, but it needs no external tools and no meta-information. 

In aggregate, one's choice of model may depend partially on what resources are available. For instance, for classic pipeline NLG systems, syntactic position and meta-information are often decided by earlier steps in the pipeline~\citep{10.5555/3241691.3241693}. Therefore, if one's aim is to rapidly construct a pipeline NLG system, then \texttt{RREG-S} should probably be preferred. 

\paragraph{Generalisability.}
We used neural REG to illustrate the importance of non-neural baselines. Our findings may not be generalisable to End2End NLG. However, if complex rule/template-based NLG systems are taken into account, \citet{dusek-etal-2018-findings} found that although these systems cannot defeat neural approaches, they still have competitive performance. It would be interesting to compare different types of models for other sub-tasks in the NLG pipeline (e.g., content determination, aggregation, and lexicalisation) in a similar way as has been done in the present paper\footnote{Note that pipelined NLG systems are sometimes thought to yield better outputs than fully End2End NLG systems~\citep{castro-ferreira-etal-2019-neural}}.

\section{Conclusion}\label{sec:conclusion}

In this work, 
we have re-evaluated state-of-the-art Neural REG systems by considering four well-designed rule- and ML-based baselines. In addition to the existing \textsc{webnlg} corpus, we built a new dataset for the task of {\em \context} on the basis of the \textsc{wsj} corpus, arguing that this dataset may be more appropriate for the task.
In the re-evaluation, we examined both our baselines and SOTA neural REG systems on both datasets, using automatic and human evaluations. The results suggest that the simplest rule-based baseline \texttt{RREG-S} achieves equally good or better performance compared to SOTA neural models. Our results on the \textsc{wsj} suggest that, on that corpus, the linguistically-informed ML-based model (\texttt{ML-L}) is best. 
We hope these results will encourage further research into the comparative strengths and weaknesses of neural, non-neural and hybrid methods in NLP.

In future, we have 4 items on our TODO list:
(1) Investigate bottleneck features for Neural based models based on the feature set of \texttt{ML-L};
(2) Explore other neural architectures (e.g., testing models that leverage pre-trained language models) and construct larger realistic REG corpora;
(3) Explore better human evaluation methods for longer documents that are better suited for evaluating the task of generating referring expressions in context;
(4) Extend our research to other languages, especially in other language families, including languages that are morphological very rich or very poor and languages that frequently use zero pronouns (e.g., Chinese~\citep{chen-etal-2018-modelling}).





\section*{Acknowledgements}
We thank the anonymous reviewers for their helpful comments.
Guanyi Chen is supported by China Scholarship Council (No.201907720022). Fahime Same is supported by the German Research Foundation (DFG)– Project-ID 281511265 – SFB 1252 “Prominence in Language” and the Junior Special Fund of the Cologne Center of Language Sciences (CCLS).
\section*{Ethical Considerations}

We collected our human evaluations using Amazon Mechanical Turk. For the \textsc{webnlg} task, which used a 7-point Likert scale, the workers were paid 0.03\$ per item, in line with rates for similar tasks. For the more demanding \textsc{wsj} task, we paid 0.10\$  per item. The payment for each task was set at \$7.5/hour (slightly above the US minimum wage, i.e., \$7.25/hour). We expected the amount to be a fair remuneration, but given the actual time some participants needed, their remuneration turned out to be on the low side. In future crowd-sourcing experiments, we will base our remuneration on a more generous estimate of the duration per experimental task.
We asked for demographic information, age, gender and English proficiency level, explicitly stating in the experiment that ``Your information will be used for research purposes only. All your data will be held \emph{anonymously}." These fields were not marked as mandatory fields. The demographic information will not be made publicly available. 

\bibliography{anthology,custom}
\bibliographystyle{acl_natbib}

\appendix




\section{\textsc{wsj} Dataset Construction} \label{sec:wsjimplementation}

In this work, we used Ontonotes 5.0 licensed by the Linguistic Data Consortium (\textsc{ldc}) \url{https://catalog.ldc.upenn.edu/LDC2013T19}. We used the files in OntoNotes Normal Form (\textsc{onf}) format. This format has combined all layers of the OntoNotes corpus, ranging from text (sentence and token segmented), parsing information, propositional content, and the coreference chains \citep{Weischedel2013}. These files were later rendered into \textsc{xml} format for full processing. As mentioned earlier, only the Wall Street Journal portion of this corpus has been used in the current study. The first and second person references were excluded from the dataset. Furthermore, we assumed that REs are presented in a linear order, therefore excluded cases such as union REs. As an example, in the sentence ``\underline{[Mary], [John] and [David]} received their booster shot last month", the REs enclosed in brackets are included in the dataset, but their \ul{underlined} union, \textit{``Mary, John, and David"}, is excluded. 
We then delexicalised the REs as explained in \S\ref{subsec:wsjdataset}. The delexicalised text in table \ref{tab:wsjsample2} shows an example from the \textsc{wsj} corpus.


\begin{table*}[ht]
	\small
	\centering

		\begin{tabular}{p{15cm}}
			\toprule
			\textbf{Text}: Ronald B. Koenig , 55 years old , was named a senior managing director of the Gruntal \& Co. brokerage subsidiary of this insurance and financial - services firm . Mr. Koenig will build the corporate - finance and investment - banking business of Gruntal , which has primarily been a retail - based firm . He  was chairman and co-chief executive officer of Ladenburg , Thalmann \& Co. until July , when he  was named co-chairman of the investment-banking firm along with Howard L. Blum Jr. , who then became the sole chief executive . Yesterday , Mr. Blum , 41 , said he  was n't aware of plans at Ladenburg  to name a co-chairman to succeed Mr. Koenig and said the board would need to approve any appointments or title changes .
			 \\ \midrule
			\textbf{Delexicialized Text}: \\
			\vspace{0.02cm}
			\textbf{Pre-context}: \ul{Mr.\_Koenig} was named a senior managing director of the \ul{Gruntal} brokerage subsidiary of this insurance and financial-services firm .
			\\
			\vspace{0.02cm}
			\textbf{Target Entity}: \ul{Mr.\_Koenig}\\
			\vspace{0.02cm}
			\textbf{Post-context}: will build the corporate-finance and investment-banking business of \ul{Gruntal}. \ul{Mr.\_Koenig} was chairman and co-chief executive officer of \ul{Ladenburg} until July , when \ul{Mr.\_Koenig} was named co-chairman of \ul{Ladenburg} along with \ul{Mr.\_Blum}. Yesterday, \ul{Mr.\_Blum} said \ul{Mr.\_Blum} was n't aware of plans at \ul{Ladenburg} to name a co-chairman to succeed \ul{Mr.\_Koenig} and said the board would need to approve any appointments or title changes.
			\\
			\bottomrule
	\end{tabular}
	\caption[An example data from the WSJ corpus.]{An example data from the WSJ corpus. In the delexicalized text, every entity is \underline{underlined}.}
	\label{tab:wsjsample2}
\end{table*}



\section{Details of \texttt{RREG-L}} \label{sec:rreg-L}
 
\begin{algorithm}[t]
    \caption{The Linguistically Informed Rule-based REG Algorithm}
    \begin{algorithmic}[1]
        \INPUT The target entity $r$, the sentence $u_2$ that $r$ is in, and its previous sentence $u_1$.
        \OUTPUT The surface form of $r$.
        \If{$r$ has an antecedent in $u_1$}
            \If{$r$ occurs in parallel context}
                \State \texttt{RealiseProRE}(r)
            \Else
                \State $\mathcal{F} \coloneqq$ \texttt{FocusSetConst}($u_1$)
                \If{$r$'s antecedent $\in \mathcal{F}$ \textbf{and} $r$ has no competitor $r' \in \mathcal{F}$}
                    \State \texttt{RealiseProRE}(r)
                \Else
                    \State \texttt{RealiseNONProRE}(r)
                \EndIf
            \EndIf
        \Else
        \State \texttt{RealiseNONProRE}(r)
        \EndIf
    \end{algorithmic}
    \label{alg:rreg}
\end{algorithm}

Algorithm~\ref{alg:rreg} describes the generation process of \texttt{RREG-L}. The system takes in the target entity $r$, the current sentence ($u_2$, i.e., the one where the target RE is located), and the previous sentence ($u_1$). 
It starts with a rule checking whether an antecedent of $r$ appears in $u_1$ (line 1). 
If the answer is no, then it realises $r$ with its non-pronominal form. 
If such an antecedent exists, the system heads to check parallelism\citep{henschel2000pronominalization}.
Concretely speaking, it checks whether or not $r$ has the same grammatical role (i.e., subject or object) as its antecedent.
If the parallelism holds, $r$ is realised as a pronoun.
Otherwise, we apply the ``local focus'' idea from~\citet{henschel2000pronominalization}, which builds upon the Centering Theory~\citep{grosz-etal-1995-centering}.
A referent is the local focus if it is (1) discourse-old, or (2) in the subject position.
In line 5, the \texttt{FocusSetConst} function constructs a set $\mathcal{F}$, consisting of local focus entities in $u_1$.
If $r$'s antecedent is an element of $\mathcal{F}$, and $r$ has no competitor being an element of $\mathcal{F}$, then we realise $r$ as a pronoun (line 6-9). 
Both surface realisation functions (i.e., \texttt{RealiseProRE} and \texttt{RealiseNONProRE}) work similarly to \texttt{RREG-S} in realising pronominal and non-pronominal REs, respectively.

\begin{table*}[htbp]
\small
\centering
\begin{tabular}{ll}
\toprule
Feature Class & Definition \\ \midrule
\texttt{Referential status} & Is $r$ the first mention of the entity in the text? \\
\texttt{Referential status} & Is the antecedent of $r$ in the same sentence? \\
\texttt{Recency} & Categorical distance between $r$ and its antecedent in number of \texttt{sentences} \\
\texttt{Recency} & Categorical distance between $r$ and its antecedent in number of \texttt{words}  \\
\texttt{Competition} & Is there any other RE between $r$ and its antecedent?  \\
\texttt{Position} & A categorical feature marking whether $r$ is the first, second, middle or last mention  \\
\bottomrule
\end{tabular}
\caption{Features used in the \texttt{ML-S} models. Each feature is defined and calculated for each target entity $r$. \texttt{Antecedent} refers to the first co-referential referring expressions preceding $r$.}
\label{tab:ml-s-feat}
\vspace{0.8cm}
\small
\centering
\begin{tabular}{ll}
\toprule
Feature Class & Definition \\ \midrule
\texttt{Grammatical Role} & Grammatical Role of $r$   \\
\texttt{Grammatical Role} &  Grammatical Role of the antecedent \\
\texttt{Meta information} & Entity type (e.g. human, city, country, organization)  \\
\texttt{Meta information} & Plurality: Is $r$ plural or singular? (only \textsc{wsj})   \\
\texttt{Meta information} & Gender \\
\texttt{Recency} & Categorical distance between $r$ and its antecedent in number of \texttt{words}\\
\texttt{Recency} & Categorical distance between $r$ and its antecedent in number of \texttt{sentences}\\
\texttt{Recency} & Categorical distance between $r$ and its antecedent in number of \texttt{paragraphs} (only \textsc{wsj})\\
\bottomrule
\end{tabular}
\caption{Features used in the \texttt{ML-L} models. }
\label{tab:ml-l-feat}
\end{table*}

\section{Detailed list of features used in \textsc{webnlg} and \textsc{wsj} feature-based ML models}\label{sec:feature}

Each feature is defined and calculated for each target entity $r$. \texttt{Antecedent} refers to the first co-referential RE preceding $r$.
Tables \ref{tab:ml-s-feat} and \ref{tab:ml-l-feat} list the features used in the ML models. 

Since we wanted to use the features in a back-off method for selecting the content of REs, we converted numerical features, such as recency, into categorical values.  We tried different recency measurements on the \textsc{webnlg} and \textsc{wsj} validation sets, and chose the ones which yielded the best results: (1) \texttt{Word distance} in all ML-based \textsc{webnlg} and \textsc{wsj} models: 5 quantile groups; (2) \texttt{Sentence distance} in \textsc{webnlg} \texttt{ML-R \& ML-S}, and \textsc{wsj} \texttt{ML-S} models: 2 quantile groups; (3) \texttt{Sentence distance} in \textsc{wsj} \texttt{ML-L}: 3 bins defined as whether the \texttt{antecedent} is in the same sentence, 1 sentence away, or more than 1 sentence away; (4) \texttt{Paragraph distance} in \textsc{wsj} \texttt{ML-L}: 4 bins defined as whether $r$ and its \texttt{antecedent} are in the same paragraph, 1 paragraph away, 2 paragraphs away, more than two paragraphs away. The paragraph information associated with the \textsc{wsj} documents are taken from the \textsc{pdtb} parser at: \url{github.com/WING-NUS/pdtb-parser/tree/master/external/aux_data/paragraphs}.

\section{Sample texts generated by the \textsc{wsj} models} \label{sec:generated_text}

 \begin{table*}[ht]
	\centering
	\begin{tabular}{p{15cm}}
	\toprule
	\textbf{\texttt{Original}}: \bl{MGM Grand Inc.} said \bl{it} filed a registration statement with the Securities and Exchange Commission for a public offering of six million common shares. \bl{The Beverly Hills , Calif.-based company} said \bl{it} would have 26.9 million common shares outstanding after the offering. \bl{The hotel and Gaming company} said Merrill Lynch Capital Markets will lead the underwriters. Proceeds from the sale will be used for remodeling and refurbishing projects , as well as for the planned MGM Grand hotel / casino and theme park.
	\\ \midrule
	\textbf{\texttt{RREG-S}}: \bl{MGM Grand Inc.} said \bl{MGM Grand Inc .} filed a registration statement with the Securities and Exchange Commission for the offering. \bl{MGM Grand Inc.} said \bl{MGM Grand Inc.} would have 26.9 million common shares outstanding after the offering. \bl{MGM Grand Inc.} said Merrill Lynch Capital Markets will lead the underwriters . Proceeds from the offering will be used for remodeling and refurbishing projects , as well as for the planned MGM Grand hotel / casino and theme park .
 \\\midrule
 \textbf{\texttt{ML-L:}} \bl{MGM Grand Inc.} said \bl{it} filed a registration statement with the Securities and Exchange Commission for an offering of common shares.
 \bl{MGM Grand Inc.} said \bl{it} would have 26.9 million common shares outstanding after the offering.
 \bl{MGM Grand Inc.} said Merrill Lynch Capital Markets will lead the underwriters . Proceeds from the offering will be used for remodeling and refurbishing projects , as well as for the planned MGM Grand hotel / casino and theme park. \\\midrule
\textbf{\texttt{ATT-Meta:}} \bl{MGM Grand Inc.} said \bl{MGM Grand Inc.} filed a registration statement with the Securities and Exchange Commission for the offering of the company of the market.
 \bl{MGM Grand Inc.} said \bl{it} would have 26.9 million common shares outstanding after the offering. \bl{MGM Grand Inc.} said Merrill Lynch Capital Markets will lead the underwriters. Proceeds from the offering will be used for remodeling and refurbishing projects , as well as for the planned MGM Grand hotel / casino and theme park .
\\
\bottomrule
	\end{tabular}
	\caption{Examples of an original text from the \textsc{wsj} dataset together with the outputs generated by the models \texttt{RREG-S}, \texttt{ML-L}, and \texttt{ATT-Meta}. References to \textit{``MGM Grand Inc.''} are boldfaced.}
	\label{tab:generatedwsj}
\end{table*}

Table \ref{tab:generatedwsj} shows a reference text from the \textsc{wsj} dataset in addition to the outputs generated by \texttt{RREG-S} (rule-based), \texttt{ML-L} (feature-based), and \texttt{ATT-Meta} (neural).

\end{document}